\documentclass[preprint,12pt,authoryear]{elsarticle}

\usepackage{graphicx}
\usepackage[fleqn]{amsmath}
\usepackage{amsthm}
\usepackage{newtxtext}
\usepackage{mathtools}
\mathtoolsset{showonlyrefs,showmanualtags}
\usepackage[varg]{newtxmath}
\usepackage{bbm}
\usepackage{xcolor}
\usepackage{subcaption}

\newtheorem{theorem}{Theorem}
\newtheorem{lemma}{Lemma}
\newtheorem{proposition}{Proposition}
\newtheorem{example}{Example}

\newcommand{\lspan}{\mathrm {span}}
\newcommand{\trc}{{\rm TR}}

\journal{}

\begin{document}

\begin{frontmatter}



\title{
Neural Tangent Kernels and
Fisher Information Matrices
for Simple ReLU
Networks with Random Hidden
Weights}


\author[inst1]{Jun'ichi Takeuchi\corref{cor1}}
\ead{tak@inf.kyushu-u.ac.jp}
\affiliation[inst1]{organization={Faculty of Information Science and Electrical Engineering, Kyushu University},%Department and Organization
            addressline={Motooka 744}, 
            city={Nishi‑ku, Fukuoka, Fukuoka},
            postcode={819‑0395}, 
            %state={State One},
            country={Japan}}

\author[inst1]{Yoshinari Takeishi}
\ead{takeishi@inf.kyushu-u.ac.jp}

\author[waseda]{Noboru Murata}
\ead{noboru.murata@eb.waseda.ac.jp}
\affiliation[waseda]{organization={Faculty of Science and  Engineering, Waseda University},%Department and Organization
            addressline={3-4-1 Okubo}, 
            city={Shinjuku-ku, Tokyo},
            postcode={169‑8555}, 
            %state={State One},
            country={Japan}}

\author[hiroshimacity]{Kazushi Mimura}
\ead{mimura@hiroshima-cu.ac.jp}
\affiliation[hiroshimacity]{organization={Faculty of Information Sciences, Hiroshima City University},%Department and Organization
            addressline={3-4-1 Ozukahigashi}, 
            city={Asaminami-ku, Hiroshima-shi, Hiroshima},
            postcode={731-3194}, 
            %state={State One},
            country={Japan}}

\author[inst12]{Ka Long Keith Ho}
\ead{ho.kalongkeith.224@s.kyushu-u.ac.jp}
\affiliation[inst12]{organization={Joint Graduate School of Math for Innovation,
Kyushu University},
%Department and Organization
            addressline={Motooka 744}, 
            city={Nishi‑ku, Fukuoka, Fukuoka},
            postcode={819‑0395}, 
            %state={State One},
            country={Japan}}

\author[uec]{Hiroshi Nagaoka}
\ead{nagaoka@is.uec.ac.jp}
\affiliation[uec]{organization={ 
The University of Electro-Communications},%Department and Organization
            addressline={1-5-1 Chofugaoka}, 
            city={Chofu-city, Tokyo},
            postcode={182-8585}, 
            %state={State One},
            country={Japan}}

\cortext[cor1]{Corresponding author}

\begin{abstract}
Fisher information matrices
and neural tangent kernels (NTK)
for 2-layer ReLU networks
with random hidden weight are argued.
We discuss the relation between both
notions as a linear transformation
and show that spectral decomposition
of NTK with concrete forms
of eigenfunctions with major
eigenvalues.
We also obtain an approximation
formula of the functions 
presented by the 
2-layer neural networks.
\end{abstract}




\end{frontmatter}


\section{Introduction}

We discuss
Fisher information matrices
and neural tangent kernels
(NTK) \citep{Jacot}
for bias-free 
2-layer Rectified Linear Unit
(ReLU) neural networks
with random hidden weights.
% The target is 2 layer (one hidden layer)
% neural networks
% with Rectified Linear Unit
% (ReLU) activation.
Assume that the input $x$ is
$d$-dimensional, the output $y$
is one-dimensional,
the hidden layer has $m$ units,
each entry of
the hidden weight matrix
$W \in \mathbb{R}^{d\times m}$
is generated according to
the Gaussian distribution with
the mean $0$ and the variance $1/m$,
$x$ is drawn from the standard
$d$-variate
Gaussian distribution denoted by $p_X$. 
The above setting is common with
\cite{TIT23,TT2024,HTT25}.

Our learning problem is to estimate
the weight vector $v \in \mathbb{R}^m$
for the output layer.
Note that the NTK is defined
as the limit that $m$ goes to infinity.
Note that we assume
$v$ is restricted as its Euclidean
norm is not more than $1$.

For this model, 
\cite{TIT23,TT2024} showed that
the eigenvalue distribution of
the Fisher information matrix of $v$
is strongly biased
and that the eigenvectors for
large eigenvalues
can be represented in terms of $W$.
It gives an approximate spectral
decomposition of the Fisher information.

In this paper, we show that
approximate spectral decomposition
for NTK holds, which is parallel to
that
for Fisher information matrix.
This result is relevant to \cite{Bietti19}.
This means that
the strong bias of eigenvalue
distribution holds for NTK as well.
Further, our decomposition
reveals the concrete expression
of eigenfuctions of NTK
with large eigenvalues.
They are
$\tilde{F}_0(x)=|x|$, 
$\tilde{F}_l(x)=x_l$ ($1 \le l \le d$),
$\tilde{F}_{\alpha\beta}(x)
=x_\alpha x_\beta /|x|$ 
($1 \le \alpha < \beta \le d$),
and $\tilde{F}'_\gamma(x)=
(x_\gamma^2 -x^2_1)/|x|$
($2 \le \gamma \le d$).
Note that $\{ \tilde{F}'_\gamma \}$
should be orthogonalized.
The eigenvalues with $\tilde{F}_0$
and with 
the group of $\tilde{F}_{\alpha\beta}$
and $\tilde{F}'_\gamma$ 
are approximately $(2d+1)/4\pi$
and $1/2\pi(d+2)$ respectively,
while the eigenvalue for 
$\tilde{F}_l$ is $1/4$.

The obtained spectral decomposition suggests
that the following approximation formula
for the function form
$f_v$ the neural network represents:
\[
f_v
\approx
f_{\tilde{\theta}}^{(D)}
= 
\sqrt{\mu_0} \, \theta_0  F_0
+\frac{1}{2}\sum_l \theta_l F_l
+\sqrt{\mu_2}
\Bigl(\sum_\gamma \theta_\gamma  F_\gamma
+\sum_{\alpha<\beta}\theta_{\alpha\beta}
F_{\alpha \beta}\Bigr),
\]
where $F_0$ etc.\ are functions
obtained by normalizing $F_0$ etc.\
with respect to the $L^2$ norm with 
respect to $p_X$,
and $\tilde{\theta}$
is the parameter consisting
of $\theta_0$, $\theta_l$, $\theta_{\alpha\beta}$
and $\theta_\gamma$.
Further,
$\mu_0$ is the eigenvalue with $F_0$,
while
$\mu_2$
is the eigenvalue with
the group of $F_\gamma$ and $F_{\alpha\beta}$.
The parameter $\tilde{\theta}$
is obtained by 
an isometric mapping of $v$
and a projection to the lower dimensional space
provided $m$ is sufficiently larger than $d^2$.
We discuss the implication 
of this approximation in Section~\ref{sec:implication}.

\section{Preliminaries}\label{sec:preliminaries}

This section provides preliminaries and problem setup. We begin by introducing ReLU neural networks in Subsection 2.1.
In Subsections 2.2 and 
2.3, we introduce
Fisher information matrices
and the notion of
neural tangent kernels (NTK), respectively.

Note that throughout this paper, all vectors are considered to be row vectors by default, unless explicitly stated otherwise.
The symbol $|v|$
denotes the Euclidean
norm of a vector $v$
and $v^T$ the transpose of $v$.

\subsection{A simple ReLU network}

We consider a simple neural network with a single hidden layer, employing the Rectified Linear Unit (ReLU) as its activation function.
The input to the network is denoted by $x \in \mathbb{R}^{1 \times d}$. 
% with its components being independent and identically distributed samples drawn from a standard normal distribution.
The output of the hidden layer is denoted by $X \in \mathbb{R}^{1 \times m}$ 
and can be represented as 
\[
    X = \phi(xW),
\]
where $\phi$ is the ReLU activation function which maps $Z \in \mathbb{R}$ to $\phi(Z) = \max\{Z, 0\}$ entry by entry. We assume that the weight matrix $W \in \mathbb{R}^{d \times m}$ is known, with its entries independently drawn from $N(0, 1/m)$,
where $N(\mu, \sigma^2)$ denotes the normal distribution with mean $\mu$ and variance $\sigma^2$. The output $y$ of the network is a scalar and is represented as
\begin{align}\label{themodel}
    y = Xv^{T} + \epsilon 
    = f_v(x) + \epsilon,
\end{align}
where $v \in \mathbb{R}^{1 \times m}$ 
is a weight vector in the last layer,
and
$\epsilon$ represents the observation noise
drawn
according to $N(0, \sigma^2)$. 
We have defined $f_v(x) = Xv^T$.
Then, the corresponding
conditional probability
density is
\[
p_v(y|x)=\frac{1}{\sqrt{2\pi\sigma^2}}\exp\Bigl(
-\frac{(y-f_v(x))^2}{2\sigma^2}
\Bigr).
\]
In this paper, 
we assume that $\sigma^2$ is a constant
and known. 
For simplicity, we assume that
$\sigma^2 = 1$ throughout the paper.
For the distribution
$p_X$
of the covariate
$x$, we assume
that $p_X$ is
the $d$-variate standard
Gaussian distribution.
Let  
$\mathcal{H}_m 
=\{ f_v : v \in \mathbb{R}^m\}$ 
and
$\mathcal{M}_m = \{ p_v : v \in \mathbb{R}^m \}$.
% In this paper, 
% we argue
% the NTK for $\{ f_v : v \in \mathbb{R}^m \}$
% and
% the Fisher information
% of $v$ in $\{ p_v : v \in \mathbb{R}^m \}$.

% \subsection{Training and generalization errors}

% Given a dataset 
% $\{(x_t, y_t) : t=1,2,\ldots,n\}$,
% we define the training error 
% % for the training dataset $
% %\{(x_t, y_t) : t=1,2,\ldots,n\}$ 
% as
% \[
% E(v)
% =\frac{1}{2n}\sum_{t=1}^n \left(y_t-
% f_v(x_t) \right)^2
% =\frac{1}{2n}\sum_{t=1}^n \left(y_t-X_t v^T\right)^2,
% \]
% where $X_t=\phi(x_tW)$. 

% Let $\hat{v}$ be the minimizer of $E(v)$ satisfying $\nabla E(\hat{v})=0$. We rewrite the training error as
% \begin{align} \label{E_v}
% E(v)=\frac{1}{2}(v-\hat{v})\hat{J}_n(v-\hat{v})^T+E(\hat{v}),
% \end{align}
% where we define
% \[
% \hat{J}_n=\frac{1}{n}\sum_{t=1}^{n} {X_t}^T X_t.
% \]
% %and $\hat{v}$ is the minimizer of $E(v)$.

% Next, we define the generalization error
% \begin{align} \label{E_star}
% E^*(v)=\frac{1}{2}\mathbb{E}\bigl[(y-Xv^T)^2\bigr]
% \end{align}
% to measure the accuracy for unseen data.
% Here, we take the expectation over the input $x$ and the noise $\epsilon$.
% Expanding the right-hand side of the above equation, we obtain
% \begin{align} \label{gen_err}
% E^*(v)=\frac{1}{2}(v-v^{*})J(v-v^{*})^T+\frac{1}{2}\sigma^2,
% \end{align}
% where $J=\mathbb{E}[X^TX]\in\mathbb{R}^{p\times p}$.
% Equation \eqref{gen_err} is minimized at $v=v^*$.
% Generally, the goal of machine learning is to minimize the generalization error $E^*(v)$ so that the developed models generalize well to new, unseen data, even though we do not have direct knowledge of the true parameter $v^*$.

\subsection{Fisher information matrix}

Let $J$ denote the Fisher information matrix
%times $\sigma^2$
. 
Then, the $(i,j)$ entry of $J$ is 
\begin{align}
J_{ij} &= \mathbb{E}\left[
\frac{\partial \log p_v(y|x)}{\partial v_i}{} 
\frac{\partial \log p_v(y|x)}{\partial v_j} \right]\\
&= 
- \mathbb{E}\left[\frac{\partial^2}{\partial v_i\partial v_j}\log p_v(y|x) \right],
\end{align}
where $v_i$ denotes the $i$th element of $v$
and $\mathbb{E}$
means the expecation with respect
to $p_v(y|x)p_X(x)$.
Since
\[
\frac{\partial^2}{\partial v_j \partial v_i}
\log p_v(y|x) 
= -\frac{X_i X_i}{\sigma^2}
= -X_i X_j,
\]
we have
\[
J
=\mathbb{E}[X^{T} X]
=\int X^TX p_X(x)dx.
\]
That is, $J$ is independent
of $v$.

Next, We define the
empirical Fisher information 
$\hat{J}$
matrix given $(x^n,y^n)$
as the Hessian of 
\[
-\frac{1}{n}\log p_v(y^n|x^n).
\]
We have
\[
\hat{J}_{ij}
= - \frac{1}{n}\sum_{t=1}^{n}\frac{\partial^2}{\partial v_i \partial v_j}\log p_v(y_t|x_t).
\]
% Hence, it holds that
% \[
% \hat{J}=\frac{1}{n}\sum_{t=1}^{n} {X_{t}}^T X_{t}
% =\hat{J}_n.
% \]

We let
$W_{i*} \in \mathbb{R}^{1 \times m}$
and
$W_{*j} \in \mathbb{R}^{d \times 1}$
denote 
the $i$th row vector of $W$ and
the $j$th column vector of $W$, respectively.
Then, we can write
\[
J_{ij}=\mathbb{E}_{x}[\phi(xW_{*i})\phi(xW_{*j}].
\]
Takeishi et al. proved the
following expansion of $J$.
\begin{align}\label{J_decom}
 J 
& = \frac{2d+1}{4\pi}v^{(0)T}v^{(0)}+\frac{1}{4}\sum_{l=1}^d W_l^T W_l \\ 
& +\frac{1}{2\pi d} \biggl(\sum_{\gamma=1}^d \!v^{(\gamma)T}v^{(\gamma)}
+\sum_{\alpha<\beta}\! 
v^{(\alpha,\beta)T}v^{(\alpha,\beta)}\!\biggr)+R,
\end{align}
where the matrix $R$ is positive-semidefinite,
and the vectors $v^{(0)}$,
$W_l$, $v^{(\gamma)}$, and
$v^{(\alpha,\beta)}$ are normalized
and approximately orthogonal
to each other
with respect to the Euclidean distance
of $\mathbb{R}^m$.

\subsection{Neural Tangent Kernel}

For
functions $f$ and $g$ over $\mathbb{R}^d$,
let $\langle f, g \rangle$ 
denote the following inner product.
\[
\langle f, g \rangle = \int f(x)g(x)p_X(x)dx,
\]
and
\[
\| f \| =\sqrt{\langle f, f \rangle}.
\]
That is,
we consider the Hilbert
space with
a usual L2 norm based on
$p_X$.

Define
\[
\mathcal{F} = \{  f : \| f \| < \infty\}.
\]
Then, $\mathcal{H}_m$ is a linear subspace of $\mathcal{F}$.

Note that 
\[
J_{ij} = 
\int X_i X_j p_X(x)dx
=
\langle X e_i^T, X e_j^T \rangle
\]
holds, 
where $e_i$ denotes the standard unit vector 
of $\mathbb{R}^m$
whose $i$th component is $1$.
We also have 
\[
\langle f_u, f_v \rangle
=\langle X u^T, X v^T \rangle
=u Jv^T.
\]

To confirm the relation of 
the Fisher information matrix
to the $L^2$ norm over $\mathcal{F}$,
consider the tangent space
of $\mathcal{M}_m$ at $p_v$.
Define the tangent space 
$\mathcal{T}_v$ by
$\mathcal{T}_v =
\lspan \{ \partial_i \}$,
where we assume
$\partial_i 
= \partial / \partial v_i$
are evaluated at $v$.
Define an
inner product for each $\mathcal{T}_v$
by
\[
\langle \partial_i, \partial_j \rangle_F
= \mathbb{E}[ \partial_i \log p_v(y|x)
\partial_j \log p_v(y|x)]
= J_{ij}.
\]
This is the Fisher metric for
$\mathcal{M}_m$.
Since $\mathcal{M}_m$ is a
linear regression model
($\mathcal{H}_m$ is a linear space)
with a fixed additive noise distribution,
the tangent space $\mathcal{T}_v$
for every $v$
with the inner product
$\langle \cdot, \cdot\rangle_F$
is isomorphic to
$\mathcal{H}_m$ with
the inner product
$\langle \cdot,\cdot \rangle$, where
$\partial_i$
corresponds to $X_i(x)$.
Further, the Fisher metric
of $\mathcal{M}_m$
is equivalent to the restriction of the $L^2$ norm
of $\mathcal{F}$ to $\mathcal{H}_m$.
In particular, 
we can regard $\mathcal{M}_m$
and $\mathcal{H}_m$
as Euclidean spaces.
Note that the distance
defined by the inner product
coincides with the square root
of doubled
Kullback-Leibler divergence.
In fact,
\begin{align}
D(p_v||p_{u})
&=\mathbb{E}_{x,y}\Bigl[
\frac{-(y-Xv^T)^2+(y-Xu^T)^2}{2} \Bigr]
\\
&=\mathbb{E}_{x,y}\Bigl[
\frac{(-Xu^T+Xv^T)(2y-Xv^T-Xu^T)}{2} \Bigr]
\\
&=\mathbb{E}_{x}\Bigl[
\frac{(-Xu^T+Xv^T)(Xv^T-Xu^T)}{2} \Bigr]
\\
&=\mathbb{E}_{x}\Bigl[
\frac{(u-v)X^TX(u-v)^T}{2} \Bigr]
\\
&=
\frac{(u-v)J(u-v)^T}{2}
\end{align}
holds.

Due to the isomorphism
mentioned above,
we can use the Fisher information
matrix as a linear operator
on $\mathcal{H}_m$
and the NTK as a linear operator
on $\mathcal{F}$ as follows.

Define the linear operator $K_m$ over $\mathcal{H}_m$
as
\[
K_m f_\theta (x) = f_{v J}(x) 
= X(x)(v J)^T. 
\]
Since 
\begin{align}
K_m f_v (x) 
& = 
X(x)Jv^T\\
&=X(x)\mathbb{E}_y[X^T(y)X(y)]v^T\\
&=
\int X(x)X^T(y)f_v(y)p_X(y)dy
\end{align}
holds,
the operator $K_m$ is extended to over 
$\mathcal{F}$,
that is, for any $f \in \mathcal{F}$, let 
\[
K_m f(x) 
=
\int X(x)X^T(y)f(y)p_X(y)dy
=
\int k_m(x,y)p_X(y)dy,
\]
where $k_m(x,y)=X(x)X^T(y)$.
Note that 
$K_m \mathcal{F} = \mathcal{H}_m$ holds.

Although $X(x)$ depends on $W$, taking the limit for 
$m \rightarrow \infty$, 
$k_m(x,y)$ %$ $\in \mathbb{R}^{\mathbb{R}}$
converges to a function over $\mathbb{R}^{d + d}$
in probability.
In fact, we have
\begin{align}
k_m(x,y)
= 
X(x)X^T(y) = \sum_{i=1}^m \phi(xW_{*i})\phi(yW_{*i})
=\frac{1}{m}
 \sum_{i=1}^m\phi(xZ_i^T)\phi(yZ_i^{T}),
\end{align}
where $Z_i = \sqrt{m}W_{*i}$.
Since each $W_{ij}$ is independently drawn from 
$N(0,1/m)$, 
each
$Z_i$ is an independent 
$d$-variate standard Gaussian variable.
Hence, by the weak law of large numbers,
$k_m(x,y)=X(x)X^T(y)$ converges in probability
to the function $k(x,y)$
defined as
\[
k(x,y)
=
\mathbb{E}_{Z_1}[\phi(xZ_1^T)\phi(yZ_1^T)].
\]
Let $K$ denote the linear operator 
over $\mathcal{F}$
given by $k(x,y)$.
% Note that 
% \[
% |\phi(xZ_1^T)\phi(yZ_1^T)|
% = |x||y|
% \phi((x/|x|)Z_1^T)\phi((y/|y|)Z_1^T)
% \le |x||y|\| Z_1 \|^2
% \]
Note that this can be though of
as
a neural tangent kernel for our model.
It induces the linear operator
$K$ over $\mathcal{F}$:
\[
K f(x) = \int k(x,y)f(y)p_X(y)dy.
\]
We must note that
the original notion of NTK
is defined based on a finite training
data set,
which is defferent from ours.
However, ours coincides with the 
original one
by letting $p_X$ be
the empirical distribution
based on the training data set.
The NTK with the empirical distribution
converges
to ours when the data size goes to
infinity.

\section{Findings}

In this section, we present our findings.
The following approximate spectral decomposition
of $K$ is shown.
\begin{theorem}\label{NTKexpansion}
The neural tangent kernel for $f_v$
is expanded as
\begin{align}\label{ntkexpansion33}
k(x,y) & =
\frac{2d+1}{4\pi }F_0(x)F_0(y)
+\frac{1}{4}\sum_{l=1}^d F_l(x)F_l(y) \\
&+\frac{1}{2\pi (d+2)}
\Biggl(\sum_{\gamma =1 }^{d-1}
F_\gamma(x)F_\gamma(y)
+\sum_{\alpha < \beta}
F_{\alpha \beta}(x)F_{\alpha \beta}(y)
\Biggr) \\
&+r(x,y),
\end{align}
where 
\begin{align}
F_0(x) & = \frac{|x|}{\sqrt{d}},\\
F_l(x) & = x_l,\\
F_\gamma(x) & 
% = \frac{\sqrt{d+2}\;
% h_\gamma(x)}{\sqrt{2}}
= \frac{\sqrt{d+2}}{\sqrt{2}}
\Biggl(
\frac{x_\gamma^2}{|x|}-\frac{|x|}{d}
-\frac{1}{\sqrt{d}+1}
\Bigl(\frac{x_d^2}{|x|}-\frac{|x|}{d}\Bigr)
\Biggr)
,\\
F_{\alpha\beta}(x) & =
\frac{\sqrt{d+2}\; x_\alpha x_\beta}{|x|},
\end{align}
and
\[
r(x,y) =
\frac{1}{2\pi}\sum_{n=1}^{\infty}\binom{2n}{n}\frac{(xy^T)^{2n+2}}{2^{2n}(2n+1)(2n+2)
(|x| |y |)^{2n+1}}.
\]
The functions $F_0$, $F_l$, $F_\gamma$,
and $F_{\alpha \beta}$ form an orthonormal system
with respect to the inner product
$\langle \cdot , \cdot \rangle$.
The remaining term $r(x,y)$ absolutely converges for any $(x,y)$.
\end{theorem}
The proof is stated in
Subsection~\ref{proof_thm_1}.

Although
$F_0$, $F_l$, $F_\gamma$,
and $F_{\alpha \beta}$ form an orthonormal system, this theorem does not claim that
they are eigenfunctions of 
$k(x,y)$. However, it is finally true.
In fact, for $F_0$ and $F_l$,
the proof is straightforward as follows.
Note that it suffices to show
$\int r(x,y) f(y)p_X(y)dx \propto f(x)$
for $f=F_0$ and $f=F_l$.
Note that for 
the cases of $f = F_0$, $F_l$, $F_\gamma$ and $F_{\alpha\beta}$,
\[
\frac{f(y)}{|y|}= f(\bar{y})
\]
holds, where $\bar{y}=y/|y|$.

We have
\begin{align}
&\int
\frac{(xy^T)^{2n+2}}{(|x||y|)^{2n+1}}
f(y)p_X(y)dy\\
= &
|x|
\int
(\bar{x}\bar{y}^T)^{2n+2}
%{\bar{y}_\alpha \bar{y}_\beta}
|y|f(y)
p_X(y)dy\\
= &
|x|
\int_{0}^{\infty}
\Bigl(
\int
(\bar{x}\bar{y}^T)^{2n+2}
\frac{f(y)}{|y|}
%{\bar{y}_\alpha \bar{y}_\beta}
\mu (d\bar{y})\Bigr)
|y|^2 p_r(|y|)d|y|\\
= &
|x|
\int_{0}^{\infty}
\Bigl(
\int
(\bar{x}\bar{y}^T)^{2n+2}
f(\bar{y})
%{\bar{y}_\alpha \bar{y}_\beta}
\mu (d\bar{y})\Bigr)
|y|^2 p_r(|y|)d|y|\\
= &
|x|
\int_{|y|=1}
(\bar{x}y^T)^{2n+2}
f(y)
\mu (dy)
\int_{0}^{\infty}r^2 p_r(r)dr,
\end{align}
where $\mu(dy)$
is the uniform probability distribution
over the 
unit sphere in $\mathbb{R}^d$
and $p_r$ is the marginal probability density of $|y|$ 
under the joint density $p_X(y)$ 
(the standard Gaussian with the dimension $d$). 
Since $\int_{0}^{\infty}r^2 p_r(r)dr$ is a constant,
our task is to evaluate
\[
\int_{|y|=1}
(\bar{x}y^T)^{2n+2}
f(y)
\mu (dy).
\]

When $f=F_0$, $f(y)$ is $1/\sqrt{d}$ under $|y|=1$.
Hence, 
\[
\int_{|y|=1}
(\bar{x}y^T)^{2n+2}
f(y)
 \mu (dy)
 =
 \int_{|y|=1}
(\bar{x}y^T)^{2n+2}d^{-1/2}
\mu (dy).
\]
Since $|\bar{x}|=1$
and since $p_\omega$ is space symmetric,
this is a constant
which does not depend on $\bar{x}$.
Hence, we have
\[
\int
\frac{(xy^T)^{2n+2}}{(|x||y|)^{2n+1}}
F_0(y)p_X(y)dy \propto |x|,
\]
where the constant of proportionality depends only on $n$.
This implies
\[
\int k(x,y)F_0(y)p_X(y)dy \propto F_0(x)
\]
and $F_0$ is an eigenfunction of the NTK $k(x,y)$.

Next, we consider $F_l(y)=y_l$.
We have
\[
\int_{|y|=1}
(\bar{x}y^T)^{2n+2}
F_l(y)
\mu (dy)
 =
 \int_{|y|=1}
(\bar{x}y^T)^{2n+2}
y_l
\mu (dy)=0,
\]
because the integrand is an odd function.
This implies $F_l$ is an eigenfunction of the NTK $k(x,y)$.
Note that
this implies 
that the eigenvalue is $1/4$.

% For $f=F_\gamma$ and $f=F_{\alpha \beta}$,
% we found an proof for 
% \[
% \int_{|y|=1}
% (\bar{x}y^T)^{2n+2}
% f(y)
%  p_\omega (dy)
%  \propto f(\bar{x}),
% \]
% but we want a more elegant proof.

% Note that 
% for $f=F_\gamma$, it suffices to show
% \[
% \int_{|y|=1}
% (\bar{x}y^T)^{2n+2}
% (y_\alpha^2-y_\beta^2)
%  p_\omega (dy)
%  \propto \bar{x}_\alpha^2-\bar{x}_\beta^2.
% \]

% \newpage

For $f=F_\gamma$ and $f=F_{\alpha \beta}$,
We can prove the 
following Theorem,
which means that
$F_\gamma$ and $F_{\alpha,\beta}$
are eigenfunctions of $k(x,y)$.

\begin{theorem}
For every $n \ge 2$
and for $f=F_\gamma$, $F_{\alpha\beta}$,
\[
\int_{|y|=1}
(\bar{x}y^T)^{2n+2}
f(y)
\mu (dy)
\propto f(x)
\]
holds.
\end{theorem}
The proof is stated in
Subsection~\ref{proof_thm_2}.

Concerning eigenfunctions
with smaller eigenvalues,
we don't have the complete
solution, but we can state
some propositions.
The following is important.
\begin{proposition}
Assume that
$f(x)$ is an eigenfunction 
with the eigenvalue $\lambda$
of the function
\[
h_n(x,y) = \frac{(xy^T)^{2n+2}}
{(|x||y|)^{2n-1}}.
\]
Then, for an aribitrary
orthogonal matrix $U$,
$f(xU)$ is an eigenfunction
of $h_n$
with the eigenvalue $\lambda$.
\end{proposition}

{\it Proof:}
By the assumption,
\[
\int 
\frac{(xy^T)^{2n+2}}
{(|x||y|)^{2n-1}}
f(y)p_X(y)dy =
\lambda f(x).
\]
Because of $p_X$'s symmetry
\[
\int 
\frac{(x(yU)^T)^{2n+2}}
{(|x||y|)^{2n-1}}
f(yU)p_X(y)dy =
\lambda f(x)
\]
holds. Hence
\[
\int 
\frac{(xU(yU)^T)^{2n+2}}
{(|x||y|)^{2n-1}}
f(yU)p_X(y)dy =
\lambda f(xU),
\]
which is
\[
\int 
\frac{(xy^T)^{2n+2}}
{(|x||y|)^{2n-1}}
f(yU)p_X(y)dy =
\lambda f(xU),
\]
{\it The proof is completed.}

By this,
we can conclude that
$(x_\alpha^2 - x_\beta^2)/|x|$
is an eigenfunction
of $k(x,y)$ with the same 
eigenvalue as
$x_\alpha x_\beta /|x|$,
even if we do not directly
prove it.

We have the following lemma.
\begin{lemma}
Define the kernel $k^{(n)}(x,y)$
for 
$n \ge 0$
as
\begin{align}\label{ntkexpansion1_finite}
k^{(n)}(x,y) &= 
\frac{|x| |y |}{2\pi}
+\frac{x y^T}{4} \\
&+\frac{1}{2\pi}\sum_{l=0}^{n}\binom{2l}{l}\frac{(xy^T)^{2l+2}}{2^{2l}(2l+1)(2l+2)
(|x| |y |)^{2l+1}}.
\end{align}
Then for $n \in \{0,1,\ldots,\lfloor (d-2)/2\rfloor \}$
\[
f(x)
=\frac{\prod_{i=1}^{2n+2}
x_{\alpha_i}}{|x|^{2n+1}}
\]
with $\alpha_1 < \alpha_2 
< \cdots < \alpha_{2n+2}$
is an eigenfunction of $k^{(n)}(x,y)$.
\end{lemma}

{\it Proof:}
Since there exists
only one term which
contains $\prod_{i=1}^{2n+2}
x_{\alpha_i}$
in the expansion of $(xy^T)^{2n+2}$
and since
the other terms in $(xy^T)^{2n+2}$
are orthogonal to $f(x)$,
$f(x)$
is an eigenfunction of
\[
\frac{(xy^T)^{2n+2}}{(|x||y|)^{2n+1}}.
\]
Further, $f(x)$
is orthogonal to $|x|$,
$x_l$, and 
\[
\frac{(xy^T)^{2l+2}}{(|x||y|)^{2l+1}}.
\]
for $l < n$.
{\it The proof is completed.}

\begin{example}
For $n=1$,
$x_\alpha x_\beta x_\gamma x_\delta
/|x|^3$
with $\alpha < \beta < \gamma < \delta$
is an eigenfunction
of $k^{(n)}(x,y)$.
Since Proposition~1,
$(x_\alpha^2-x_\beta^2)x_\gamma x_\delta/|x|^3$
and
$(x_\alpha^2-x_\beta^2)(x_\gamma^2
-x_\delta^2)/|x|^3$
are eigenfuctions with
the same eigenvalue
as that of
$x_\alpha x_\beta x_\gamma x_\delta
/|x|^3$.
\end{example}

Let $\lambda_i$ is the $i$th eigenvalue of 
the linear operator $K$.
Then, we have
\begin{align}
\sum_{i=1}^\infty \lambda_i 
= &
\int k(x,x)p_X(x)dx = \mathbb{E}_{x,Z}[\phi(xZ^T)\phi(xZ^T)]\\
= & \mathbb{E}_{x,Z}[(\phi(xZ^T))^2]
= \mathbb{E}_{x,Z}[|x|^2(\phi(\bar{x}Z^T))^2]\\
= & \mathbb{E}_{x,V}[|x|^2(\phi(V))^2]=
\frac{\mathbb{E}_{x}[|x|^2]}{2}=
\frac{d}{2},
\end{align}
where $V$ is a standard Gaussian 
random variable and $\bar{x}$ denotes $x/|x|$.
Hence,
\begin{align}
\int r(x,x)p_X(x)dx
& \le \frac{d}{2}
-\frac{2d+1}{4\pi }
-\frac{d}{4}
-\frac{(d-1)+d(d-1)/2}{2\pi (d+2)}\\
&
=\frac{d}{2}\Bigl(\frac{1}{2}
-\frac{(2d+1)(d+2)+2(d-1)+d(d-1)}{2\pi d(d+2)} 
\Bigr)\\
&
=\frac{d}{2}\Bigl(\frac{1}{2}
-\frac{3d^2+2d}{2\pi d(d+2)} \Bigr)\\
&
=\frac{d}{2}\Bigl(\frac{1}{2}
-\frac{3d+2}{2\pi (d+2)}\Bigr),
\end{align}
Note that
\[
\frac{1}{2}
-\frac{3d+2}{2\pi (d+2)}
=
\frac{1}{2}
-\frac{3d+6}{2\pi (d+2)}
+\frac{4}{2\pi (d+2)}
=
\frac{1}{2} \Bigl(
1- \frac{3}{\pi}
\Bigr)
+\frac{2}{\pi (d+2)}
\]
holds.
Since $1-3/\pi \le 1 -3/3.142 
\le 0.0452$,
when $d$ is sufficiently large,
the trace $\trc(r(x,y))$
is less than $0.026$
$\times \trc(k(x,y))$.

We can think of 
Theorem~\ref{NTKexpansion} 
as an approximate spectral decomposition
of $k(x,y)$.
% For example, $F_0$ is approximately equal
% to an
% eigenfunction of the first eigenvalue.
% This is parallel to 
% the approximate spectral decomposition
% of Fisher information matrix 
% \eqref{J_decom}.
In the successive section, we
discuss the relation 
of the approximate spectral decomposition
of $k(x,y)$
with that for the Fisher information
matrix.

% suggests that
% we can obtain a 
% spectral decomposition of 
% $k(x,y)$ 
% by continuing the process of orthogonalization
% for higher order terms in $r(x,y)$,

\section{Relation Between
Spectral Decomposition of NTK and 
That of Fisher
Information}

Since the Fisher information $J$ is defined for
the finite dimensional model 
$\mathcal{M}_m$,
first we discuss the relation
between $J$ and $k_m(x,y)$.

Assume that $J$ has the following 
spectral decomposition:
\begin{align}\label{J_assumed_decom}
J = \sum_{i=1}^m \lambda_i u^{(i)T}u^{(i)},
\end{align}
where 
$\{ u^{(i)} \}$  
(each element belongs to $\mathbb{R}^{1 \times m}$) is
an orthonormal basis of $\mathbb{R}^m$
with respect to the standard inner product 
$u^{(i)} u^{(j)T}$.
% \[
% \langle u^{(i)}, u^{(j)}\rangle_s = u^{(i)} u^{(j)T}.
% \]
Note that 
the projectors $u^{(i)T}u^{(i)}$,
which forms the spectral decomposition,
is with respect to the 
standard inner product.

Noting that $\sum_{i=1}^m u^{(i)T} u^{(i)}$
equals the identity matrix of order $m$,
we have
\begin{align}
k_m(x,y)
=
X(x)X(y)^T
=
X(x)\Bigl(\sum_{i=1}^m u^{(i)T} u^{(i)} \Bigr) X(y)^T
% = 
% \phi(xW)
% \Bigl(\sum_{i=1}^m u^{(i)T} u^{(i)} \Bigr) 
% \phi(yW)^T
=
\sum_{i=1}^m f_{u^{(i)}}(x)f_{u^{(i)}}(y).
\end{align}
Recall that $k_m$ is used as
$K_m f=\int k_m(x,y)f(y)p_X(y)dy$
to obtain the image of the operator $K_m$.
This process is related to 
the projection with respect to 
the $L^2$ norm with the kernel $p_X$.
In particular,
each $f_{u^{(i)}}(x)f_{u^{(i)}}(y)$
is
a constant times a projector
with respect to the $L^2$ norm.
In fact, since
\[
\| f_{u^{(i)}} \|
= \sqrt{u^{(i)}Ju^{(i)T}}
=\sqrt{\lambda_i},
\]
we can rewrite it as 
\[
f_{u^{(i)}}(x)f_{u^{(i)}}(y)
= 
\lambda_i \tilde{f}_{u^{(i)}}(x)
\tilde{f}_{u^{(i)}}(y),
\]
where $\tilde{f}_{u^{(i)}}(x)=
f_{u^{(i)}}/\sqrt{\lambda_i}$
and 
$\{ f_{u^{(i)}}/\sqrt{\lambda_i} \}$
is an orthonormal basis of 
$\mathcal{H}_m$
with respect to the $L^2$ norm.
That is, we have 
the follwoing spectral decomposition
with respect to the $L^2$ norm:
\begin{align}\label{K_assumed_decom}
k_m(x,y)
=
\sum_{i=1}^m \lambda_i 
\tilde{f}_{u^{(i)}}(x)\tilde{f}_{u^{(i)}}(y),
\end{align}
% where $\tilde{f}_{u^{(i)}}(x)$
% denotes $f_{u^{(i)}}(x)/\sqrt{\lambda_i}$,
% that is, $\|\tilde{f}_{u^{(i)}} \|=1$.
This is also represented as
\begin{align}\label{decom_ntk_L2}
K_m f = \sum_{i=1}^m \lambda_i 
\tilde{f}_{u^{(i)}}
\langle \tilde{f}_{u^{(i)}}, f \rangle.
\end{align}
Via this form,
we construct
another spectral decomposition
in terms of the different inner product
$(\cdot,\cdot)$ defined as
\[
( f, g ) = 
\langle f, K_m^{-1} g\rangle
=
\int f(x) 
\Bigl( \int k_m^{-1}(x,y)g(y)
p_X(y)dy\Bigr)
p_X(x)dx,
\]
where
\[
k_m^{-1}(x,y)=
\sum_{i=1}^m \frac{1}{\lambda_i} 
\tilde{f}_{u^{(i)}}(x)\tilde{f}_{u^{(i)}}(y),
\]
Since
\[
K^{-1}_m \tilde{f}_{u^{(i)}} 
=\frac{1}{\lambda_i} 
\tilde{f}_{u^{(i)}}
\]
% \[
% (f_{u^{(i)}},f_{u^{(j)}})
% =
% \langle 
% f_{u^{(i)}},K^{-1}_mf_{u^{(j)}} \rangle
% =\frac{1}{\lambda_j} 
% \langle 
% f_{u^{(i)}},f_{u^{(j)}} \rangle
% =\frac{\lambda_i \delta_{ij}}{\lambda_j}
% =\delta_{ij}
% \]
holds, we have
$(f_{u^{(i)}},f_{u^{(j)}})
=\langle f_{u^{(i)}},
K^{-1}_mf_{u^{(j)}} \rangle
=
\langle f_{u^{(i)}},
f_{u^{(j)}} \rangle/\lambda_j 
=\delta_{ij}$
and
\begin{align}
K_m f 
 = 
\sum_{i=1}^m \lambda_i 
\tilde{f}_{u^{(i)}}
\langle 
\lambda_i
K^{-1}_m
\tilde{f}_{u^{(i)}}, f \rangle
= 
\sum_{i=1}^m \lambda_i^2
\tilde{f}_{u^{(i)}}
(
\tilde{f}_{u^{(i)}}, f ).
% = 
% \sum_{i=1}^m \lambda_i
% f_{u^{(i)}}
% (
% f_{u^{(i)}}, f ).
% \mbox{ and }
% (f_{u^{(i)}},f_{u^{(j)}})
% & =\delta_{ij},
\end{align}
which yields
\begin{align}
\label{NTK_assumed_decom_2}
K_m f 
 = 
\sum_{i=1}^m \lambda_i
f_{u^{(i)}}
(
f_{u^{(i)}}, f ).
\end{align}
This is the spectral decomposiion
of $K_m$ with repect to
the inner product $(\cdot,\cdot)$.
% since $(f_{u^{(i)}},f_{u^{(i)}})=1$.
Note that
$(f_{u^{(i)}},f_{u^{(j)}})
=\delta_{ij}$
cooresponds to
$u^{(i)}u^{(j)T}=\delta_{ij}$,
which implies
$(f_{u},f_{v})=uv^T$.
Hence, 
the counterpart
of \eqref{J_assumed_decom}
is \eqref{NTK_assumed_decom_2},
not \eqref{decom_ntk_L2}.

The fact $(f_{u},f_{v})=uv^T$
should be remarked.
Define another parameter 
$\theta$ than $v$ by
\[
f_v =
g_\theta
=\sum_{i=1}^m
\theta_i f_{v^{(i)}}.
\]
Then, $\langle g_\theta,g_{\theta'} \rangle 
= \sum_{i=1}^m \lambda_i\theta_i \theta_i'$ 
and $(g_\theta,g_{\theta'}) 
= \theta \theta'^T$
hold.
That is, $\theta$ is the parameter
which diagonalizes $J$.
Hence, 
$\theta = v U$ holds
with a certain orthogonal matrix.
It means that the gradient
by $\theta$
is equivalent to that by $v$. 

Next, we argue the situation 
in which $m \rightarrow \infty$.
From Theorem~\ref{NTKexpansion}
and the fact that
the trace of $k(x,y)$ is $d/2$,
the NTK has
a discreet spectral decomposition.
Hence, assume that we have
\[
k(x,y) = \sum_{i=1}^\infty
\lambda_i \tilde{g}_i(x)\tilde{g}_i(y),
\]
with $\| \tilde{g}_i \| = 1$
and $\sum_{i}\lambda_i = d/2$.
Since the image of $K$ is 
$\mathcal{H}=\bigcup_m K_m
\mathcal{F}$,
$\{ \tilde{g}_i \}$ is an orthonormal 
basis of $\mathcal{H}$ 
with respect to the $L^2$ norm.
Similarly as the finite dimensional case,
let 
$g_i = \sqrt{\lambda_i}\tilde{g}_i$
and
\[
g_\theta = \sum_{i=1}^\infty
\theta_i g_i.
\]
Note that
\[
\mathcal{H}
= 
\Bigl\{
g_\theta 
=
\sum_{i}\theta_i \sqrt{\lambda_i} \tilde{g}_i
: \sum_{i} \lambda_i \theta^2_i < \infty
\Bigr\}.
\]
holds.
Define the inner product $(\cdot,\cdot)$ over $\mathcal{H}$
as
\[
(g_\theta,g_{\theta'})
=\theta \theta'^{T}
=\sum_{i=1}^\infty \theta_i \theta'_i.
\] 
Then, the spectral decomposition
of $k(x,y)$ based on $(\cdot, \cdot)$
is
\[
K g = \sum_i \lambda_i g_i (g_i,g).
\]

\section{Implication of Spectral Decomposition
of NTK}
\label{sec:implication}

Let $p_\theta(y|x)$ denote
the conditional probability
density defined by $g_\theta$. 
Then, we have
\begin{align}\label{KLexpansion}
D(p_{\hat{\theta}}||p_\theta)
=
\frac{\| g_{\hat{\theta}} 
- g_{\theta}\|^2}{2}
= \frac{1}{2}\sum_{i=1}^\infty 
\lambda_i(\hat{\theta}_i -\theta_i)^2.
\end{align}
Hence, 
\[
\frac{\partial D(p_{\hat{\theta}}||p_\theta)}{\partial \theta_i}
= -\lambda_i(\hat{\theta}_i 
-\theta_i).
\]
holds.
Now define gradient
$\nabla$ with respect to 
$\theta \in \mathbb{R}^{\mathbb{N}}$ as
\[
\nabla = \sum_{i=1}^\infty
\frac{\partial}{\partial \theta_i}.
\]
Then, we have
\[
\nabla D(p_{\hat{\theta}}||p_\theta) 
= - K[g_{\hat{\theta}}-g_{\theta}],
\]
which derives a learning dynamics
related to the neural tangent kernel.
This is a consequence
of the fact that
$\theta$ for the finite $m$ case
is obtained by
an isometric mapping of $v$.

We can remove the parameterization
$\theta$ by employing an alternative
definition of $\nabla$ as
\[
\nabla D(\hat{p}||p) =
\arg \max_{q:(q,q) = 1}
D(\hat{p}||p+q).
\]

The expansion \eqref{KLexpansion}
suggests the following approximation
to $g_\theta$:
\begin{align}\label{aaa}
g_\theta(x)
=
\sum_{i=1}^D \theta_i g_i(x)
+r_D(x)
=
f_{\tilde{\theta}}^{(D)}(x)
+r_D(x)
,
\end{align}
where 
\begin{align}
r_D(x) &= \sum_{i=D+1}^\infty
\theta_i g_i(x),\\
f_{\tilde{\theta}}^{(D)}(x)
&=\sum_{i=1}^D \theta_i g_i(x).
\end{align}
Recall that each $g_i$
is normalized with respect to
the inner product $(\cdot,\cdot)$,
which means
$\| g_i \| = \sqrt{\lambda_i}$.
Since we assume that
$\sum_{i=1}^\infty \theta_i^2$
is not more than $1$,
the squared $L^2$ norm
of $r_D$ is not more than
\[
\sum_{i=D+1}^\infty \lambda_i,
\]
which converges to zero as $D$
goes to infinity.
Hence, if we use
\eqref{aaa} to approximate
${p_v}$ with very large $m$,
the estimation error for
$f_{\tilde{\theta}}^{(D)}$
and the maximum $L_2$ norm
of $r_D$ correspond to
the estimation error
and
the approximation error
in the model selection to choose
the optimal $D$.

For example,
if we set $D$ so that it
covers up to $F_{\alpha\beta}$
and $F_\gamma$,  
the squared $L^2$ norm
of $r_D$ is not more than
$d/2$ times 0.026.
Although we do not know their
norms based on $(\cdot,\cdot)$
except for $F_l$, 
we showed that $F_0$, $F_l$, $F_{\gamma}$,
and $F_{\alpha\beta}$ are eigenfunctions
of $k(x,y)$.
Hence, our approximation model is
\[
f_{\tilde{\theta}}^{(D)}
= 
\sqrt{\mu_0} \, \theta_0  F_0
+\frac{1}{2}\sum_l \theta_l F_l
+\sqrt{\mu_2}
\Bigl(\sum_\gamma \theta_\gamma  F_\gamma
+\sum_{\alpha<\beta}\theta_{\alpha\beta}
F_{\alpha \beta}\Bigr),
\]
where
$\mu_0$ is the eigenvalue
with $F_o$
and
$\mu_2$ 
that with $F_\gamma$ and $F_{\alpha\beta}$.
The parameter $\tilde{\theta}$
is the vector
consisting of $\theta_0$,
$\theta_l$ ($1\le l \le d$),
$\theta_\gamma$
($1\le \gamma \le d-1$),
and
$\theta_{\alpha\beta}$
($\alpha < \beta$).
For the eigenvalues, we have
\[
\frac{2d+1}{4\pi }
\le
\mu_0 
\le \frac{2d+1}{4\pi }
+\frac{0.026 d}{2}
\]
and
\[
\frac{1}{2\pi (d+2)}
\le
\mu_2 \le 
\frac{1}{2\pi (d+2)}
+
\frac{0.026 d}{2}\cdot \frac{2}{d(d+3)}.
\]
Important is that
the range of parameter
is
$|\tilde{\theta}| \le 1$,
which corresponds to
$|v| \le 1$ for finite $m$ cases.
Hence,
the factors
$\sqrt{\mu_0}$,
$1/2$,
and $\sqrt{\mu_2}$
represent
the importance of
each of $F_0$, $F_l$, $F_\gamma$,
and $F_{\alpha\beta}$ in
the training process.
Furthermore,
we should note that
the sample size to obtain
sufficient accuracy for each
parameter is proportional to
the inverse of the corresponding eigenvalue.
Since the eigenfunction with
smaller eigenvalue is more complex,
this observation seems reasonable.

Further, our observation here
seems relevant to the classic
theory of inductive inference
which claims that learning without bias
is impossible.
Here, the bias is called
prior distributions, prior knowledge,
penalty (in penalized maximum likelihood),
etc. in various learning strategies.
In this context, 
the 2-layer networks are quiped with
the natural bias that simpler
terms are preferred.

% Then, we have
% \[
% \langle f_i, f_i \rangle_a = \lambda^{-1}.
% \]
% Let $\tilde{f}_i = f_i \sqrt{\lambda_i}$.
% Then, $\langle \tilde{f}_i, \tilde{f}_i \rangle_a =1$.

\section{The proof of main results}

We show the proofs of the main results of 
this paper.

\subsection{Proof of Theorem~\ref{NTKexpansion}}
\label{proof_thm_1}

Consider the relation of 
\[
k(x,y)
=\mathbb{E}_{Z_1}[\phi(xZ_1^T)\phi(yZ_1^T)].
\]
to
\[
J_{ij}=\mathbb{E}_{x}[\phi(xW_{*i})\phi(xW_{*j})].
\]
In fact, by exchanging the roles of
$(x,y,Z_1)$ and $(W_{*i},W_{*j},x)$, the NTK
$k(x,y)$ becomes to $J_{ij}$.

Referring the proof of Lemma~2 of \cite{TIT23,TT2024},
we have for $J_{ij}$ with the fixed $m$,
\begin{align*}
J_{ij} &= 
% \frac{A_{ij}}{2\pi}\left(1+\sum_{n=0}^{\infty}\binom{2n}{n}
% \frac{(W^{(i)} \cdot \!W^{(j)})^{2n+2}}{2^{2n}(2n+1)(2n+2)A_{ij}^{2n+2}}\right)\\
% & +\frac{W^{(i)}\! \cdot \!W^{(j)}}{4}\\
% &
\frac{A_{ij}}{2\pi}
+\frac{W^{(i)}\! \cdot \!W^{(j)}}{4}
+\frac{(W^{(i)}\!\cdot \!W^{(j)})^2}{4\pi A_{ij}}\\
&+\frac{1}{2\pi}\sum_{n=1}^{\infty}\binom{2n}{n}\frac{(W^{(i)}\!\cdot \!W^{(j)})^{2n+2}}{2^{2n}(2n+1)(2n+2)A_{ij}^{2n+1}},
\end{align*}
where $A_{ij} = | W_{*i}| |W_{*j} |$.
This implies that we have
\begin{align}\label{ntkexpansion1}
k(x,y) &= 
\frac{|x| |y |}{2\pi}
+\frac{x y^T}{4}
+\frac{(xy^T)^2}{4\pi |x| |y |} \\
&+\frac{1}{2\pi}\sum_{n=1}^{\infty}\binom{2n}{n}\frac{(xy^T)^{2n+2}}{2^{2n}(2n+1)(2n+2)
(|x| |y |)^{2n+1}}.
\end{align}
This expansion absolutely converges for any $(x,y)$.
Using this expansion, we will find
an approximate spectral decomposition of $k(x,y)$.
For the second and third terms, we have
\begin{align}
\frac{xy^T}{4}&=\sum_{i}\frac{x_iy_i}{4}\\
    \frac{(xy^T)^2}{4\pi|x| |y |}
    &=
    \frac{\sum_{ij}x_iy_i x_j y_j}{4\pi|x||y |}
    =
    \frac{\sum_{ij}x_ix_jy_iy_j}{4\pi|x| |y|}
    =
    \frac{\sum_{i}x^2_iy^2_i}
    {4\pi|x| |y |}
     +
    \frac{2\sum_{i < j}x_ix_jy_iy_j}
    {4\pi|x| |y |}.
\end{align}
% Here, 
% $|x|$, $x_l$,
% $x_\gamma^2/|x|$, and $x_\alpha x_\beta/|x|$
% are linearly independent,
% where $1 \le l \le d$,
% $1\le \gamma \le d$, and 
% $1 \le \alpha < \beta \le d$.

From the above expansion, we see
that  
$|x|$, $x_l$ ($1\le l \le d$),
$x_\gamma^2/|x|$, and $x_\alpha x_\beta/|x|$
($1 \le \alpha < \beta \le d$)
are eigenvectors of $k(x,y)$,
if they are orthogonal to each other
and if we ignore the last term of \eqref{ntkexpansion1}.
In fact, $\lspan\{|x|\}$,
$\lspan\{ x_l \}$, 
$\lspan\{ x_\gamma^2/|x| \}$,
and
$\lspan\{ x_\alpha x_\beta /|x|\}$
are orthogonal to each other
except for the pair of
$\lspan\{ |x| \}$ and $\lspan\{x_\gamma^2 /|x| \}$, where 
$\lspan\{ x_l \}$ is the linear span
of the set of $x_l$ for $1 \le l \le d$ 
for example.

The orthogonality follows 
from the fact that the products of
the pairs of functions, one from
one span and another from another span,
are odd in an $x_k$ with a certain $k$;
$|x|\cdot x_l$, 
$|x|\cdot x_\alpha x_\beta/|x|
=x_\alpha x_\beta$,
%$x_l\cdot x_{l'}$ ($l \neq l'$),
$x_l \cdot x_\gamma^2/|x|$, 
$x_l \cdot x_\alpha x_\beta/|x|$, and
$x_\gamma^2/|x|\cdot x_\alpha x_\beta/|x|
=x_\gamma^2x_\alpha x_\beta/|x|^2$.

% As for each linear span,
% its basis functions 
% (e,g. $x_l$'s for $\lspan\{ x_l\}$)
% are orthogonal to each other except
% for the pair
% $\lspan\{ |x|\}$ and
% $\lspan\{ x_\gamma^2/|x| \}$.
% It follows from the oddness of the 
% products. For example
% $x_l x_{l'}$ with $l \neq l'$ is odd.

Concerning $\lspan\{ |x| \}$
and $\lspan\{ x_\gamma^2  /|x|\}$,
we will orthogonalize their basis
$|x|$ and $x_\gamma^2 /|x|$'s below.

Note that they are linearly dependent
because
\[
\sum_{\gamma =1}^d \frac{x_\gamma^2}{|x|} = |x|.
\]
Hence, we will remove $x_d^2/|x|$
in the final form.

Now we modify $\sum_{i=1}^d x^2_iy^2_i$ as
\begin{align}
   & \sum_{i=1}^d x^2_iy^2_i\\
= &
\sum_{i=1}^d\Biggl(
\Bigl(x^2_i -\frac{|x |^2}{d}\Bigr)
\Bigl(y^2_i - \frac{|y |^2}{d} \Bigr)
+
\frac{y_i^2|x |^2}{d}
+
\frac{x_i^2|y |^2}{d}
-
\frac{|x |^2 |y |^2}{d^2}
\Biggr)\\
= &
\sum_{i=1}^d
\Bigl(x^2_i -\frac{|x |^2}{d}\Bigr)
\Bigl(y^2_i - \frac{|y |^2}{d} \Bigr)
+
\frac{|y |^2 |x |^2}{d}
+
\frac{| x|^2 |y |^2}{d}
-
\frac{|x |^2 |y |^2}{d}\\
= &
\sum_{i=1}^d
\Bigl(x^2_i -\frac{|x |^2}{d}\Bigr)
\Bigl(y^2_i - \frac{|y |^2}{d} \Bigr)
+
\frac{|x |^2 |y |^2}{d}.
\end{align}
Hence, we have
\begin{align}
 \frac{|x| |y |}{2\pi} 
+ \frac{\sum_{i}x^2_iy^2_i}
    {4\pi|x| |y |}
    = &
       \frac{(2d+1)|x| |y |}{4\pi d}
 +\frac{1}{4\pi}
\sum_{i=1}^d
\Bigl(\frac{x^2_i}{|x|} -\frac{|x |}{d}\Bigr)
\Bigl(\frac{y^2_i}{|y|} - \frac{|y |}{d} \Bigr)\\
= &
\frac{2d+1}{4\pi d}g_0(x)g_0(y)
+\frac{1}{4\pi}\sum_{\gamma =1 }^d
g_\gamma(x)g_\gamma(y),
\end{align}
where we have defined
$g_0(x)=|x|$ and 
\[
g_\gamma(x)
=\frac{x_\gamma^2}{|x|}-\frac{|x|}{d}.
\]
We can confirm the orthogonality of $g_0$
and $g_\gamma$ as
\[
\langle g_0,g_\gamma \rangle
=
\mathbb{E}\Bigl[|x|\frac{x_\gamma^2}{|x|}
-\frac{|x|^2}{d}\Bigr]
=1-\frac{d}{d} = 0.
\]

Finally, we will find an orthogonal basis
for $\lspan\{ g_\gamma : 1 \le \gamma \le d\}$.
First, we note that
\begin{align}
\| g_\gamma \|^2 
& =
\mathbb{E}\Bigl[ \Bigl(\frac{x^2_\gamma}{|x|} -\frac{|x |}{d}\Bigr)^2 \Bigr]\\
&=
\mathbb{E}\Bigl[ \frac{x^4_\gamma}{|x|^2} 
-\frac{2x_\gamma^2 }{d}
+\frac{|x |^2}{d^2} \Bigr]\\
&=\frac{3}{d+2}-\frac{2}{d} + \frac{1}{d}\\
&=\frac{2d-2}{d(d+2)}.
\end{align}
% ($\mathbb{E}[ x_\gamma^ 4/|x|^2]
% = 3/(d+2)$ was shown by Takeishi-kun.)

Note that $g_\gamma$'s
are the nodes of a regular $(d-1)$ dimensional simplex in $\lspan\{ g_\gamma \}$, 
whose center of gravity is the origin of
$\lspan\{ g_\gamma \}$.
Hence, we have
$\langle g_\gamma, g_{\gamma'}\rangle$
$= C_g \| g_\gamma \|^2 
(\delta_{\gamma \gamma'}-1/d)$,
where $C_g = (1-1/d)^{-1}$
Let $h_\gamma = g_\gamma + r g_d$.
Then, it is easy to confirm
\[
\langle h_\gamma, h_{\gamma'} \rangle
= C_g\|g_\gamma \|^2\delta_{\gamma \gamma'}
\]
for $1 \le \gamma, \gamma' \le d-1$,
if $r = -1/(\sqrt{d}+1)$
or $r = 1/(\sqrt{d}-1)$.
We employ the former, i.e.\ we let
\[
h_\gamma = g_\gamma 
- \frac{g_d}{\sqrt{d}+1}.
\]
Note that 
\[
\| h_\gamma \| 
= \sqrt{C_g}\| g_\gamma\|
= \frac{\sqrt{d}}{\sqrt{d-1}}
\frac{\sqrt{2d-2}}{\sqrt{d(d+2)}}
=
\frac{\sqrt{2}}{\sqrt{d+2}}
\]
holds.
We have
\begin{align}
\sum_{\gamma=1}^{d-1}
h_\gamma(x)h_\gamma(y)
& =
\sum_{\gamma=1}^{d-1}
\Bigl(g_\gamma(x)
-\frac{g_d(x)}{\sqrt{d}+1}\Bigr)
\Bigl(g_\gamma(y)
-\frac{g_d(y)}{\sqrt{d}+1}\Bigr)\\
& =
\sum_{\gamma=1}^{d-1}
g_\gamma(x)g_\gamma(y) 
+ \frac{g_d(x)g_d(y)}{\sqrt{d}+1}
 + \frac{g_d(x)g_d(y)}{\sqrt{d}+1}\\
& + \frac{(d-1)g_d(x)g_d(y)}{(\sqrt{d}+1)^2}\\
& =
\sum_{\gamma=1}^{d-1}
g_\gamma(x)g_\gamma(y) 
+ \frac{2g_d(x)g_d(y)}{\sqrt{d}+1}
 + \frac{(\sqrt{d}-1)g_d(x)g_d(y)}{\sqrt{d}+1}\\
 &= \sum_{\gamma=1}^{d}
g_\gamma(x)g_\gamma(y).
\end{align}
Hence, we have
\begin{align}\label{ntkexpansion2}
k(x,y) & =
\frac{2d+1}{4\pi d}|x||y|
+\frac{1}{4}\sum_{l=1}^d x_ly_l
\\
&+\frac{1}{4\pi}
\Biggl(
\sum_{\gamma =1 }^{d-1}
h_\gamma(x)h_\gamma(y)
+\sum_{\alpha < \beta}
 \frac{\sqrt{2}x_\alpha x_\beta }{|x| }
 \frac{\sqrt{2} y_\alpha y_\beta}{|y |}
\Biggr)
 \\
&+\frac{1}{2\pi}\sum_{n=1}^{\infty}\binom{2n}{n}\frac{(xy^T)^{2n+2}}{2^{2n}(2n+1)(2n+2)
(|x| |y |)^{2n+1}}.
\end{align}

Next, we examine the norms of $|x|$, $x_l$, $h_\gamma$, etc.
We have
\begin{align}
\| |x|\|&=
\bigl(\mathbb{E}[|x|^2]\bigr)^{1/2}=\sqrt{d},\\
\| x_l\|&=
\bigl(\mathbb{E}[x_l^2]\bigr)^{1/2}=1,\\
\| h_\gamma \| & = 
\frac{\sqrt{2}}{\sqrt{d+2}},\\
\mbox{ and } \:
\Biggl\| \frac{\sqrt{2}x_\alpha x_\beta}{|x|}  \Biggr\|&=
\Biggl(\mathbb{E}\Biggl[\frac{2 \, x^2_\alpha x^2_\beta}{|x|^2}
\Biggr]\Biggr)^{1/2}=\frac{\sqrt{2}}{\sqrt{d+2}}.
\end{align}
We define the normalized vectors:
\begin{align}
F_0(x) & = \frac{|x|}{\sqrt{d}},\\
F_l(x) & = x_l,\\
F_\gamma(x) & = \frac{\sqrt{d+2}\;
h_\gamma(x)}{\sqrt{2}}
= \frac{\sqrt{d+2}}{\sqrt{2}}
\Biggl(
\frac{x_\gamma^2}{|x|}-\frac{|x|}{d}
-\frac{1}{\sqrt{d}+1}
\Bigl(\frac{x_d^2}{|x|}-\frac{|x|}{d}\Bigr)
\Biggr)
,\\
\mbox{ and } \:
F_{\alpha\beta}(x) & =
\frac{\sqrt{d+2}\; x_\alpha x_\beta}{|x|}. 
\end{align}
We have
\begin{align}\label{ntkexpansion3}
k(x,y) & =
\frac{2d+1}{4\pi }F_0(x)F_0(y)
+\frac{1}{4}\sum_{l=1}^d F_l(x)F_l(y)
\\
&+\frac{1}{2\pi (d+2)}
\Biggl(\sum_{\gamma =1 }^{d-1}
F_\gamma(x)F_\gamma(y)
+\sum_{\alpha < \beta}
F_{\alpha \beta}(x)F_{\alpha \beta}(y)
\Biggr)
 \\
&+r(x,y),
\end{align}
where we have defined
\[
r(x,y) =
\frac{1}{2\pi}\sum_{n=1}^{\infty}\binom{2n}{n}\frac{(xy^T)^{2n+2}}{2^{2n}(2n+1)(2n+2)
(|x| |y |)^{2n+1}}.
\]
The last term $r(x,y)$ is shown to be positive definite
owing to Schur's product theorem.

\subsection{Proof of Theorem~2}
\label{proof_thm_2}

First consider $F_{\alpha \beta}$.
Let $e_\alpha$ and $e_\beta$ denote
the basis vectors for
the coordinates 
$x_\alpha$ and $x_\beta$
respectively.
Decompose $\bar{x}$ to
the directions of 
$\lspan \{e_\alpha,e_\beta \}$
and its orthogonal complement as
\[
\bar{x} = \bar{x}_{\rm p}
+\bar{x}_{\rm n},
\]
where $\bar{x}_{\rm p} 
= \bar{x}_\alpha e_\alpha
+\bar{x}_\beta e_\beta$.
Let $e_{\rm n}$ denote
the unit vector to the direction of 
$\bar{x}_n$.
%  and let $\bar{x}_{\rm n} 
% = e_{\rm n}\bar{x}$.
% Then
% \[
% \bar{x}=
% \bar{x}_{\rm p} e_{\rm p}
% +\bar{x}_\alpha e_\alpha
% +\bar{x}_\beta e_\beta.
% \]
Then, $\{e_\alpha, e_\beta, e_{\rm n} \}$ is an orthonormal system
and $\bar{x}$ belongs to
$V=\lspan \{ e_\alpha, e_\beta, e_{\rm n} \}$.
Let $V^{\perp}$ denote
the orthogonal complement of $V$.
Let $z$ denote
the projection of $y$ to 
$\lspan \{ e_\alpha, e_\beta, e_{\rm n} \}$
and let $w = y-z$. 
Then, $\bar{x}y^T = \bar{x}z^T$ holds.

Now, the target integral 
is reduced to
\[
I = \int_{|y|=1}
(\bar{x}z^T)^{2n+2}
z_\alpha z_\beta
\mu (dy).
\]
Introduce the new
variables for integration
by
\begin{align}
z &= \cos \varphi \cdot
\bar{z} \\
w &= \sin \varphi \cdot
\bar{w}
\end{align}
where $\bar{z}$
belongs to the
unit sphere in $V$,
$\bar{w}$ belongs to 
the unit sphere in 
$V^{\perp}$,
and $0 \le \varphi \le \pi/2$.
Then, we have
\[
I = \int_0^{\pi/2}
\cos^{2n+6} \varphi
\int
(\bar{x}\bar{z}^{T})^{2n+2}
\bar{z}_\alpha \bar{z}_\beta
\mu^{(2)}(d\bar{z})
\sin^{d-3} \varphi
\int
\mu^{(d-3)} (d\bar{w})d\varphi,
\]
where $\mu^{(d-1)}$
denotes the uniform
probability measure
over the
unit sphere in $\mathbb{R}^d$.
We have
\[
I = 
\int
(\bar{x}\bar{z}^{T})^{2n+2}
\bar{z}_\alpha \bar{z}_\beta
\mu^{(2)}(d\bar{z})
\int_0^{\pi/2}
\cos^{2n+6} \varphi
\sin^{d-3} \varphi d\varphi.
\]
Since the integral
with respect to
$\varphi$ is independent of $\bar{x}$,
our task is to evaluate
\[
I_2 = 
\int
(\bar{x}\bar{z}^{T})^{2n+2}
\bar{z}_\alpha \bar{z}_\beta
\mu^{(2)}(d\bar{z}).
\]

For the evaluation of $I_2$,
consider the conditional expectation
given $\bar{x}\bar{z}^T$.
Assume that
$\bar{z}$ is restricted as
$\bar{x}\bar{z}^T = s$.
Then, 
$\bar{z}$
is uniformly distributed along
the circle of the radius $\sqrt{1-s^2}$
centered at $s\bar{x}$, in the
plane orthogonal to $\bar{x}$.
Then, $\bar{z}_\alpha e_\alpha + \bar{z}_\beta e_\beta$
is uniformly distributed
along the ellipse in 
$\lspan\{ e_\alpha, e_\beta \}$
which
is the projection of the circle
where $\bar{z}$ exists.
The center of the ellipse
is
$s \bar{x}_\alpha e_\alpha 
+ s\bar{x}_\beta e_\beta$,
which contributes
$s^{2n+2}
s^2 x_\alpha x_\beta$
to the conditional expectation.
We will evaluate
the conditional expectation
with the distribution on
the congruent ellipse
centered at the origin.
Note that the direction
of its minor axis is along with
$\bar{x}_\alpha e_\alpha
+ \bar{x}_\beta e_\beta$,
its minor radius
is 
\[
\sqrt{1-s^2} \sqrt{1-\bar{x}_\alpha^2-\bar{x}_\beta^2},
\]
and its major radius is
$\sqrt{1-s^2}$.
Hence, the ellipse is
represented with the parameter $\theta$
as
\begin{align}
(\bar{z}_\alpha,\bar{z}_\beta)^T
=\sqrt{1-s^2}
U(\varphi)
(\gamma \cos \theta,\sin \theta)^T
,
\end{align}
where $\gamma=\sqrt{1-x_\alpha^2-x_\beta^2}$,
$(\cos \varphi,\sin \varphi)
= (\bar{x}_\alpha,\bar{x}_\beta)
/|(\bar{x}_\alpha,\bar{x}_\beta)|
$,
and $U(\varphi)$ is
the rotation matrix of angle $\varphi$. We have
\begin{align}
\bar{z}_\alpha 
&=\sqrt{1-s^2}
(\cos \varphi \cdot \gamma \cos \theta-\sin
\varphi \sin \theta)\\
\bar{z}_\beta
&= \sqrt{1-s^2}
(\sin \varphi \cdot \gamma \cos \theta
+ \cos \varphi \sin \theta),
\end{align}
which yields
\[
\bar{z}_\alpha \bar{z}_\beta
=(1-s^2)
\bigl(
\cos \varphi \sin \varphi
(\gamma^2 \cos^2 \theta 
-\sin^2 \theta)
+\gamma
(\cos^2 \varphi-\sin^2 \varphi)
\cos \theta \sin \theta
\bigr).
\]
Hence, the conditional expectation
of $\bar{x}_\alpha \bar{x}_\beta$
is
\[
\frac{1}{2\pi}
\int_0^{2\pi} \bar{z}_\alpha \bar{z}_\beta d\theta
=\frac{1}{2}(1-s^2)(\gamma^2-1)
\sin \varphi \cos \varphi
= -\frac{(1-s^2)\bar{x}_\alpha 
\bar{x}_\beta}{2},
\]
where $\gamma^2-1 
= -|(\bar{x}_\alpha,\bar{x}_\beta)|^2$
is used.
Hence, with the contribution
from the center of the ellipse,
we have the following.
\[
I_2 = \mathbb{E}_s
\Bigl[
s^{2n+4}\bar{x}_\alpha\bar{x}_\beta
-\frac{s^{2n+2}(1-s^2)\bar{x}_\alpha\bar{x}_\beta}{2}
\Bigr]  \propto \bar{x}_\alpha \bar{x}_\beta.
\]
As for $F_{\gamma}$,
it suffices to examine
$g_\alpha(x)
=
\bar{x}_\alpha^2 - \bar{x}_1^2$,
since each $F_\gamma$
is a linear combination of $g_\alpha$
($2 \le \alpha \le d$).
This task can be done similarly
as $\bar{x}_\alpha \bar{x}_\beta$.
{\it The proof is completed.}

%\section{Conclusion}

\section*{Acknowledgment}
This work was supported by JSPS KAKENHI Grant Number JP23H05492.
% MIRAI2.0 – Joint seed funding of Japan-Sweden 
% collaborative projects, and the Sasakawa foundation.

\appendix

\bibliographystyle{elsarticle-harv} 
\bibliography{cas-refs}


\end{document}